\documentclass{article}
\usepackage{times}

\usepackage{multicol}
\usepackage[bookmarks=true]{hyperref}

\usepackage{perpage}
\MakePerPage{footnote}
\usepackage{eqparbox}
\usepackage[pdftex]{graphicx}
\usepackage{enumitem}
\usepackage{todonotes}
\usepackage{arydshln}
\usepackage{amsmath}
\usepackage{amsfonts} 
\usepackage{subcaption}
\usepackage{tabularx}
\usepackage{float}
\usepackage{wrapfig}
\usepackage{sidecap}
\sidecaptionvpos{figure}{t}
\usepackage{algorithm}
\usepackage{algorithmic}
\usepackage{authblk}

\newcommand{\target}{{T}}
\newcommand{\att}{{h}}
\newcommand{\attparam}{{\psi}}
\newcommand{\actparam}{{\theta}}
\newcommand{\qparam}{{\phi}}
\newcommand{\tmodule}{{A_\target}}
\newcommand{\smodule}{{A_s}}
\newcommand{\omodule}{{A_o}}
\newcommand{\segmap}{{z}}
\newcommand{\objatt}{{c}}

\newcommand{\method}{Attention-Privileged Reinforcement Learning}
\newcommand{\met}{APRiL}

\usepackage{authblk}
\usepackage[final]{corl_2020}
\title{\method}

\author{\textbf{Sasha Salter$^1$, Dushyant Rao$^2$, Markus Wulfmeier$^2$, Raia Hadsell$^2$, Ingmar Posner$^1$} \\
$^1$Applied AI Lab, University of Oxford,  \texttt{\{sasha, ingmar\}@robots.ox.ac.uk} \\

$^2$Deepmind, London, \texttt{\{dushyantr, mwulfmeier, raia\}@google.com}
}

\begin{document}
\maketitle

%===============================================================================

\begin{abstract}
Image-based Reinforcement Learning is known to suffer from poor sample efficiency and generalisation to unseen visuals such as distractors (task-independent aspects of the observation space). Visual domain randomisation encourages transfer by training over visual factors of variation that may be encountered in the target domain. This increases learning complexity, can negatively impact learning rate and performance, and requires knowledge of potential variations during deployment. In this paper, we introduce \method~(\met) which uses a self-supervised attention mechanism to significantly alleviate these drawbacks: by focusing on task-relevant aspects of the observations, attention provides robustness to distractors as well as significantly increased learning efficiency. \met~trains two attention-augmented actor-critic agents: one purely based on image observations, available across training and transfer domains; and one with access to privileged information (such as environment states) available only during training. Experience is shared between both agents and their attention mechanisms are aligned. The image-based policy can then be deployed without access to privileged information. We experimentally demonstrate accelerated and more robust learning on a diverse set of domains, leading to improved final performance for environments both within and outside the training distribution \footnotemark[3] \footnotetext[3]{Videos comparing \met~and asym-DDPG baseline: \\ https://sites.google.com/view/april-domain-randomisation/home}. 
\end{abstract}


%===============================================================================
\section{Introduction}

While image-based Deep Reinforcement Learning (RL) has recently provided significant successes in various high-data domains \citep{mnih2015human,silver2017mastering,lillicrap2015continuous}, its application to physical systems remains challenging due to expensive and slow data generation, challenges with respect to safety, and the need to be robust to unexpected changes in the environment.

When training visual models in simulation, we can obtain robustness either by adaptation to target domains~\citep{ganin2016domain, bousmalis2016domain,  wulfmeier2017mutual}, or by randomising system parameters with the aim of covering all possible environment parameter changes~\citep{tobin2017domain, rajeswaran2016epopt, openai2018handmanip, pinto2017asymmetric, sadeghi2016cad2rl, viereck2017learning}. Unfortunately, training under a distribution of randomised visuals ~\citep{sadeghi2016cad2rl, viereck2017learning}, can be substantially more difficult due to the increased variability. This often leads to a compromise in final performance~\citep{openai2018handmanip, tobin2017domain}. Furthermore, it is usually not possible to cover all potential environmental variations during training. Enabling agents to generalise to unseen visuals such as distractors (task-independent aspects of the observation space) is an important question in robotics where an agent's environment is often noisy (e.g. autonomous vehicles). 

To increase robustness and reduce training time, we can  make use of privileged information such as environment states, commonly accessible in simulators. By using lower-dimensional, more structured and informative representations directly as agent input, instead of noisy observations affected by visual randomisation, we can improve data efficiency and generalisation~\citep{tassa2018deepmind,peng2018sim}.

However, raw observations can be easier to obtain and dependence on privileged information during deployment can be restrictive. When exact states are available during training but not deployment, we can make use of information asymmetric actor-critic methods \citep{pinto2017asymmetric, schwab2019simultaneously} to train the critic faster via access to the state while providing only images for the actor.

\begin{figure*}
    \centering
    \includegraphics[width=0.98\linewidth]{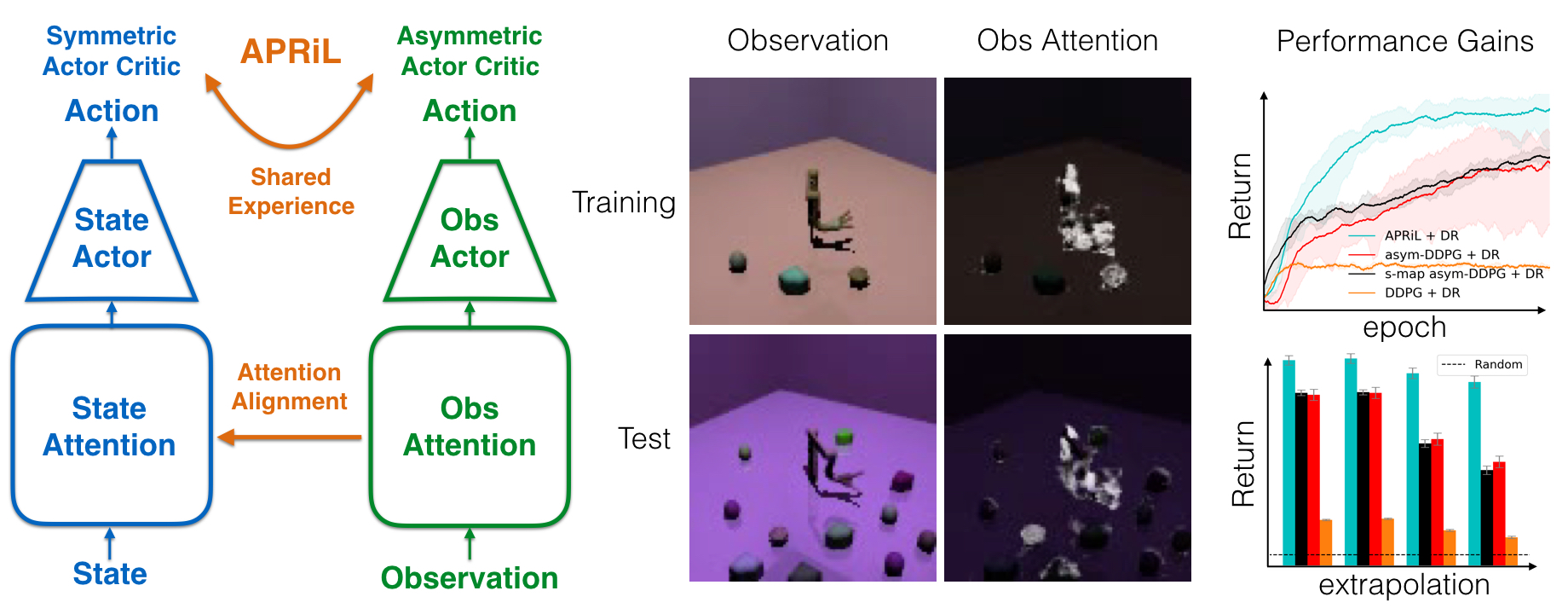} \&
    \caption{\textbf{Model diagram (left):} \met~concurrently trains two attention augmented policies (one state-based, the other image-based). \textbf{Qualitative and quantitative results (middle \& right):} By aligning the observation attention to that of the state, image-based attention quickly suppresses highly varying, task-irrelevant, information \textbf{(middle second column)}. This leads to increased learning rate \textbf{(top right)} and robustness to extrapolated domains with increasing levels of unseen additional distractors \textbf{(bottom right)}. For \textit{JacoReach}, attention (\textbf{middle second column}; white and black signify high and low values) is paid only to the target object and jaco arm in training and extrapolated domains.}
    \label{fig:SARL_simple_model}
\end{figure*}

By introducing \method~(\met), we further leverage privileged information readily and commonly available during training, such as simulator states and object segmentations~\citep{todorov2012mujoco, dosovitskiy2017carla}), for increased robustness, sample efficiency, and generalisation to distractors. As a general extension to asymmetric actor-critic methods, \met~concurrently trains two actor-critic systems (one symmetric with a state-based agent, the other asymmetric with an image-dependent actor).
Both actors utilise attention to filter their inputs, and we encourage alignment between both attention mechanisms. As state-space learning is unaffected by visual randomisation, the observation attention module efficiently attends to state- and task-dependent aspects of the image whilst explicitly becoming invariant to task-irrelevant and noisy aspects of the environment (distractors). We demonstrate that this leads to faster image-based policy learning and increased robustness to task-irrelevant factors (both within and outside the training distribution). See Figure \ref{fig:SARL_simple_model} for a visualisation of \met~, its attention, and generalisation capabilities on one of our domains.

In addition, \met~shares a replay buffer between both agents, which further accelerates training for the image-based policy. At test-time, the image-based policy can be deployed without privileged information. We test our approach on a diverse set of simulated domains across robotic manipulation, locomotion, and navigation; and demonstrate considerable performance improvements compared to competitive baselines when evaluating on environments from the training distribution as well as in extrapolated and unseen settings with additional distractors.

\section{Problem Formulation}
\label{background}

Before introducing \method~(\met), this section provides a 
background for the RL algorithms used. For a more in-depth introduction please refer to~\citet{lillicrap2015continuous} and~\citet{pinto2017asymmetric}.

\subsection{Reinforcement Learning}

We describe an agent's environment as a Partially Observable Markov Decision Process which is represented as the tuple ${(S, O, A, P, r, \gamma, s_0)}$, where $S$ denotes a set of continuous states, $A$ denotes a set of either discrete or continuous actions, $P: S \times A \times S \rightarrow \{ x \in \mathbb{R} | 0 \le x \le 1\}$ is the transition probability function, $r : S \times A \rightarrow \mathbb{R}$ is the reward function, $\gamma$ is the discount factor, and $s_0$ is the initial state distribution. $O$ is a set of continuous observations corresponding to continuous states in $S$. At every time-step $t$, the agent takes action $a_t = \pi(\cdot|s_t)$ according to its policy $\pi : S \rightarrow A$. The policy is optimised as to maximize the expected return $R_t = E_{s_0}[\sum_{i=t}^{\infty}\gamma^{i-t}r_i|s_0]$. The agent's Q-function is defined as $Q_\pi(s_t, a_t) = E[R_t|s_t, a_t]$.

\subsection{Asymmetric Deep Deterministic Policy Gradients}
\label{sec:addpg}

\begin{wrapfigure}{r}{0.45\textwidth}
    \centering
    \vspace{-20pt}
    \includegraphics[width=0.98\linewidth]{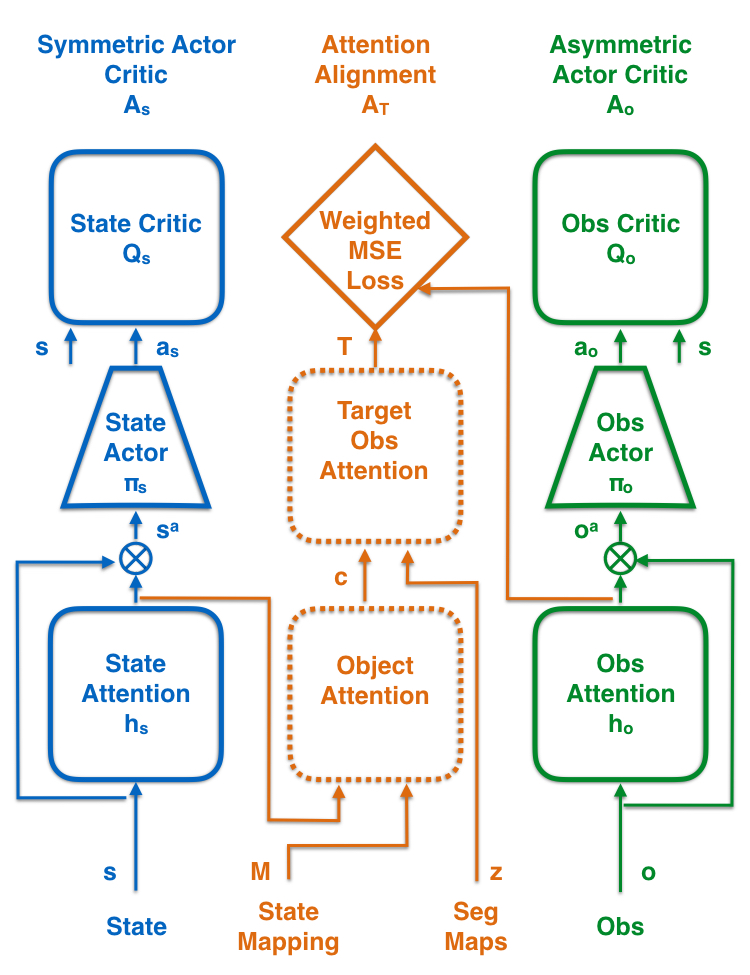}
    \caption{\textbf{\met~'s architecture}. Blue, green and orange represent symmetric and asymmetric actor critic and attention alignment modules ($A_s$, $A_o$, $A_T$). The diamond represents the attention alignment loss. Dashed and solid blocks are non-trainable and trainable networks. The $\otimes$ operator signifies element-wise multiplication. Experiences are shared using a shared replay buffer.}
    \vspace{-60pt}
    \label{fig:SARL_MODEL}
\end{wrapfigure}

Asymmetric Deep Deterministic Policy Gradients (asymmetric DDPG)~\citep{pinto2017asymmetric} represents a type of actor-critic algorithm designed specifically for efficient learning of a deterministic, observation-based policy in simulation. This is achieved by leveraging access to more compressed, informative environment states, available in simulation, to speed up and stabilise training of the critic.

The algorithm maintains two neural networks: an observation-based actor or policy $\pi_\actparam : O \rightarrow A$ (with parameters $\actparam$) used during training and test time, and a state-based Q-function (also known as critic) $Q^\pi_\qparam : S \times A \rightarrow R$ (with parameters $\qparam$) which is only used during training. 

To enable exploration, the method (like its symmetric version~\citep{silver2014deterministic}) relies on a noisy version of the policy (called behavioural policy), e.g. $\pi_b(o) = \pi(o) + z$ where $z \sim \mathcal{N}(0, 1)$ (see Appendix \ref{hyperparams} for our particular instantiation). The transition tuples $(s_t, o_t, a_t, r_t, s_{t+1}, o_{t+1})$ encountered during training are stored in a replay buffer~\citep{mnih2015human}. Training examples sampled from the replay buffer are used to optimize the critic and actor. By minimizing the Bellman error loss $\mathcal{L}_{critic} = (Q(s_t, a_t) - y_t)^2$, where $y_t = r_t+\gamma Q(s_{t+1}, \pi(o_{t+1}))$, the critic is optimized to approximate the true Q values. The actor is optimized by minimizing the loss: \newline $\mathcal{L}_{actor} = - E_{s, o \sim \pi_b(o)}[Q(s, \pi(o))]$.

\section{\method~(\met)}

\met~improves the robustness and sample efficiency of an observation-based agent by using multiple ways to benefit from privileged information. First, we use an asymmetric actor-critic setup~\citep{pinto2017asymmetric} to train the observation based actor. Second, we additionally train a quicker learning state-based actor, while sharing replay buffers, and aligning attention mechanisms between both actors. We emphasise here that our approach can be applied to any asymmetric, off-policy, actor-critic method~\citep{konda2000actor} with the expectation of similar performance benefits to those demonstrated in this paper. Specifically we choose to build off Asymmetric DDPG~\citep{pinto2017asymmetric} due to its accessibility. 

\met~is comprised of three modules as displayed in Figure \ref{fig:SARL_MODEL}. The first two modules, $\smodule$ and $\omodule$, each represent a separate actor-critic with an attention network incorporated over the input for each actor. For the \textit{state-based} module $\smodule$ we use standard symmetric DDPG, while the \textit{observation-based} module $\omodule$ builds on asymmetric DDPG, with the critic having access to states. Finally, the third part $\tmodule$ represents the alignment process between attention mechanisms of both actor-critic agents to more effectively transfer knowledge between the both learners respectively. 

\textbf{$\smodule$} consists of three networks: $Q^\pi_s$, $\pi_s$, $\att_s$ (critic, actor, and attention) with parameters $\{{\qparam_s}, {\actparam_s}, {\attparam_s}\}$. Given input state $s_t$, the attention network outputs a soft gating mask $\att_t$ of same dimensionality as the input, with values ranging between $[0,1]$. The input to the actor is an attention-filtered version of the state, $s^a_t = \att_s(s_t) \odot s_t$. To encourage a sparse masking function, we found that training this attention module on both the traditional DDPG loss as well as an entropy loss helped:
\begin{equation}
  \mathcal{L}_{h_s} = - E_{s \sim \pi_b}[Q_s(s, \pi_s(s^a)) - \beta H(\att_s(s))],
  \label{eq: actor_state_loss}
\end{equation}
    \noindent where $\beta$ is a hyperparameter (set through grid-search, see Appendix \ref{hyperparams}) to weigh the additional entropy objective, and $\pi_b$ is the behaviour policy that obtained experience (in this case from a shared replay buffer). The actor and critic networks $\pi_s$ and $Q_s$ are trained with the symmetric DDPG actor and Bellman error losses. We found that \met~was not sensitive to the absolute value of $\beta$, only the magnitude, and was set low enough to not suppress task-relevant parts of the state-space.

Within \textbf{$\tmodule$}, the state-attention obtained in $\smodule$ is converted to corresponding observation-attention $\target$ to act as a self-supervised target for the observation attention module in $\omodule$.
This is achieved in a two-step process. First, state-attention $\att_s(s)$ is converted into object-attention $c$, which specifies how task-relevant each object in the scene is. The procedure uses information about which dimension of the environment state relates to which object. Second, object-attention is converted to observation-space attention by performing a weighted sum over object-specific segmentation maps \footnote{Simulators (e.g., \citep{todorov2012mujoco, dosovitskiy2017carla}) commonly provide functionality to access these segmentations and semantic information for the environment state.}:
\begin{equation}
    \objatt = M \cdot \att_s(s), \qquad \target =  \sum_{i=0}^{N - 1} \objatt_i \cdot \segmap_i
    ~~~~\label{eqn_3}
\end{equation} 

Here, $M \in \{0,1\}^{N \times n_s}$ ($n_s$ is the dimensionality of $s$) is an environment-specific, predefined adjacency matrix that maps the dimensions of $s$ to each corresponding object, and $\objatt \in [0,1]^N$ is an attention vector over the $N$ objects in the environment. $\objatt_i$ corresponds to the $i^{th}$ object attention value. $\segmap_i \in \{0,1\}^{W \times H}$ is the binary segmentation map of the $i^{th}$ object segmenting the object with the rest of the scene, and has the same dimensions as the image. $\segmap_i$ assigns values of $1$ for pixels in the image occupied by the $i^{th}$ object, and $0$ elsewhere. $T \in [0,1]^{W \times H}$ is the converted state-attention to observation-space attention to act as a target on which to train the observation-attention network $\att_o$.
\begin{algorithm}[h!]
  \caption{\method}
  \label{alg:example}
\begin{algorithmic}
  \STATE Initialize the actor-critic modules $\smodule$, $\omodule$, attention alignment module $A_T$, replay buffer $R$
  \FOR{episode$=1$ {\bfseries to} $M$}
  \STATE Initial state $s_0$
  \STATE Set DONE $\leftarrow$ FALSE
  \WHILE{$\neg$ DONE}
  \STATE Render image observation $o_t$ and segmentation maps $z_t$:
  \STATE \hskip4.5em  $o_t, z_t \leftarrow$ {renderer}$(s_t)$
  \IF{episode \textbf{mod} $2 = 0$}
  \STATE Obtain action $a_t$ using obs-behavioral policy and obs-attention network:
  \STATE \hskip4.5em  $a_t \leftarrow \pi_{o}(h_o(o_t) \odot o_t)$
  \ELSE
  \STATE Obtain action $a_t$ using state-behavioral policy and state-attention network:
  \STATE \hskip4.5em  $a_t \leftarrow \pi_{s}(h_s(s_t) \odot s_t)$
  \ENDIF
  \STATE Execute action $a_t$, receive reward $r_t$, DONE flag, and transition to $s_{t+1}$
  \STATE Store $(s_t, o_t, z_t, a_t, r_t, s_{t+1}, o_{t+1})$ in $R$
  \ENDWHILE
  \FOR{$n=1$ {\bfseries to} $N$}
  \STATE Sample minibatch $\{s, o, z, a, r, s^{'}, o^{'}\}_0^B$ from $R$
  \STATE Optimise state- critic, actor, and attention using $\{s, a, r, s^{'}\}_0^B$ with $\smodule$
  \STATE Convert state-attention to target observation-attention $\{T\}_0^B$ using $\{s, o, z\}_0^B$ with $A_T$
  \STATE Optimise observation- critic, actor, and attention using $\{s, o, T, a, r, s^{'}, o^{'}\}_0^B$ with $\omodule$
  \ENDFOR
  \ENDFOR
\end{algorithmic}
\label{algo:APRiL}
\end{algorithm}
The observation module $\omodule$ also consists of three networks: $Q^\pi_o$, $\pi_o$, $\att_o$ (respectively critic, actor, and attention) with parameters $\{{\phi_o},{\actparam_o}, {\attparam_o}\}$. The structure of this module is the same as $\smodule$ except the actor and critic now have asymmetric inputs. The actor's input is the attention-filtered version of the observation, $o^a_t = \att_o(o_t) \odot o_t$ \footnote{In practice, the output of $\att_o(o_t)$ is tiled to match the number of channels that the image contains}. The actor and critic $\pi_o$ and $Q_o$ are trained with the asymmetric DDPG actor and Bellman error losses in \ref{sec:addpg}.
The main difference between $\omodule$ and $\smodule$ is that the observation attention network $\att_o$ is trained on both the actor loss and an object-weighted mean squared error loss:
\begin{equation}
\mathcal{L}_{h_o} = E_{o, s \sim \pi_b} [\frac{1}{2}\sum_{ij}\frac{1}{w_{ij}}(\att_o(o) - T)^2_{ij}  - \nu Q_o(s, \pi_o(o^a))]
\end{equation}

where weights $w_{ij}$ denote the fraction of the image $o$ that the object present in $o_{i,j,1:3}$ occupies, and $\nu$ represents a hyperparameter for the relative weighting of both loss components (see Appendix \ref{hyperparams} for exact value). The weight terms, $w$, ensure that the attention network becomes invariant to the size of objects during training and does not simply fit to the most predominant object in the scene.

During training, experiences are collected evenly from both state and observation based agents and stored in a shared replay buffer (similar to~\citet{schwab2019simultaneously}). This is to ensure that: 1. Both state-based critic $Q_s$ and observation-based critic $Q_o$ observe states that would be visited by either of their respective policies. 2. The attention modules $h_s$ and $h_o$ are trained on the same data distribution to better facilitate alignment. 3. Efficient discovery of highly performing states from $\pi_s$ are used to speed up learning of $\pi_o$.

Algorithm \ref{algo:APRiL} shows the pseudocode for a single actor implementation of \met. In practice, in order to speed up data collection and gradient computation, we parallelise the agents and environments and ensure equal data is generated by state- and image-based agents.

\section{Experiments}

\begin{figure*}
    \includegraphics[width=0.32\linewidth]{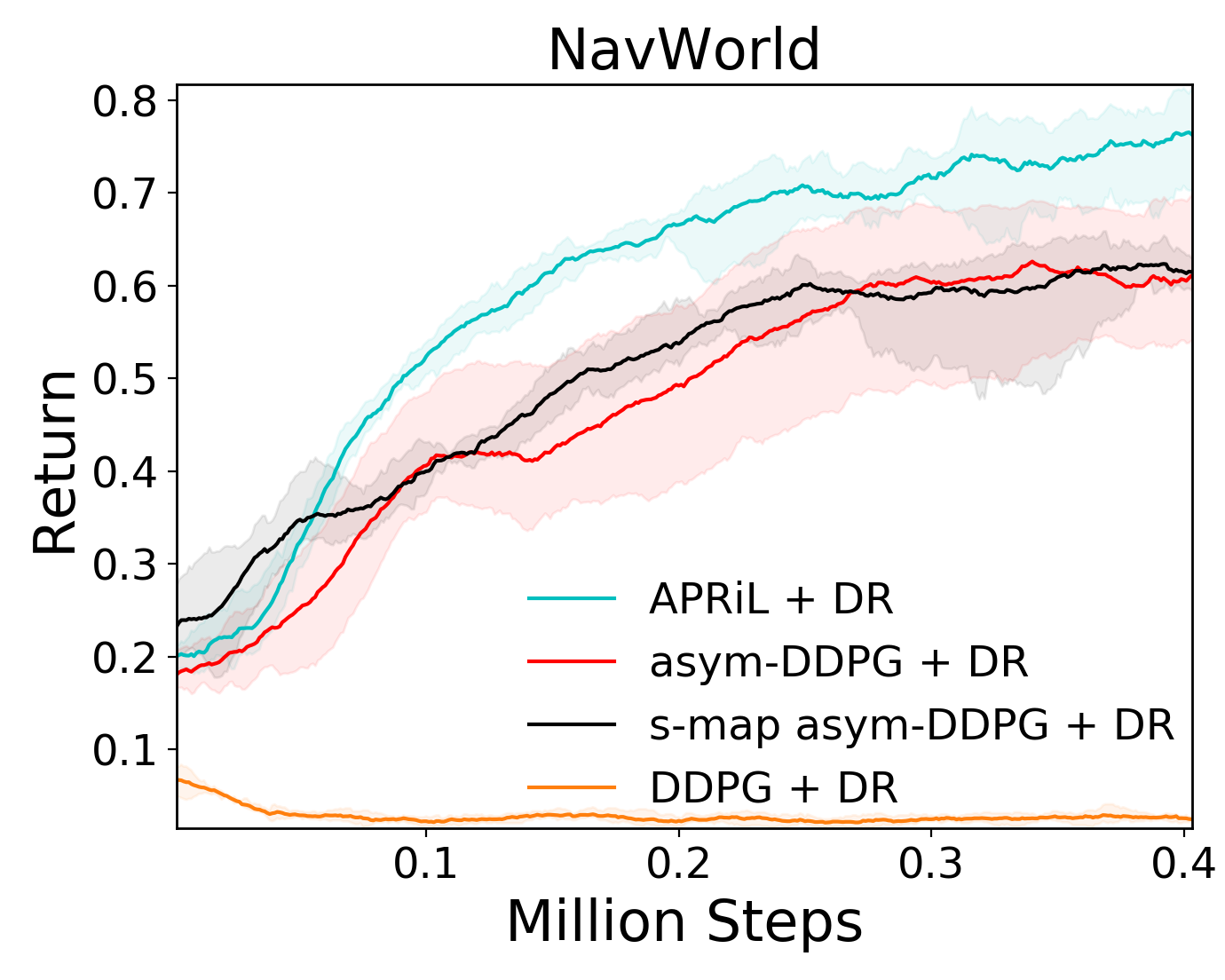}
    \includegraphics[width=0.32\linewidth]{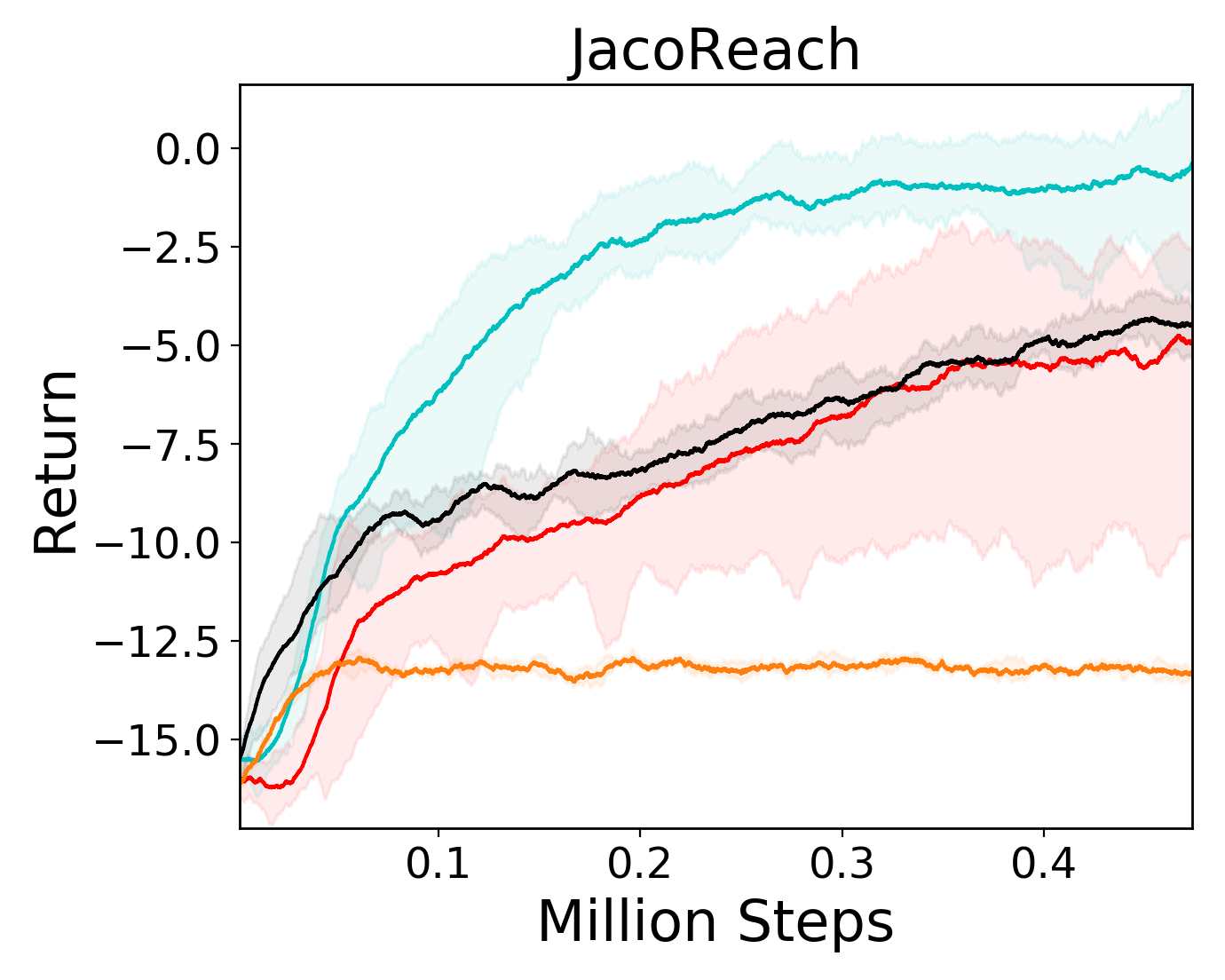}
    \includegraphics[width=0.32\linewidth]{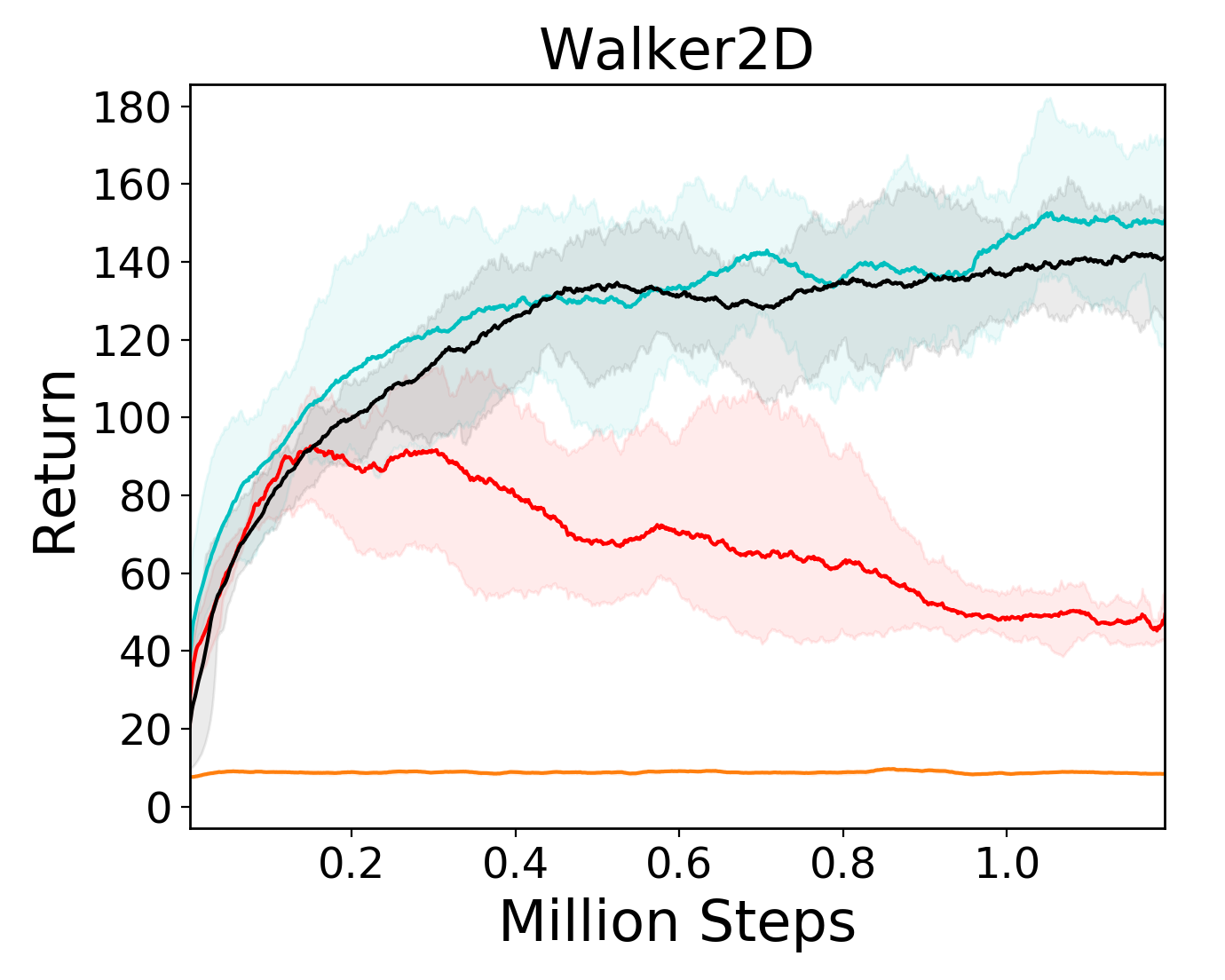} \\
    \includegraphics[width=0.32\linewidth]{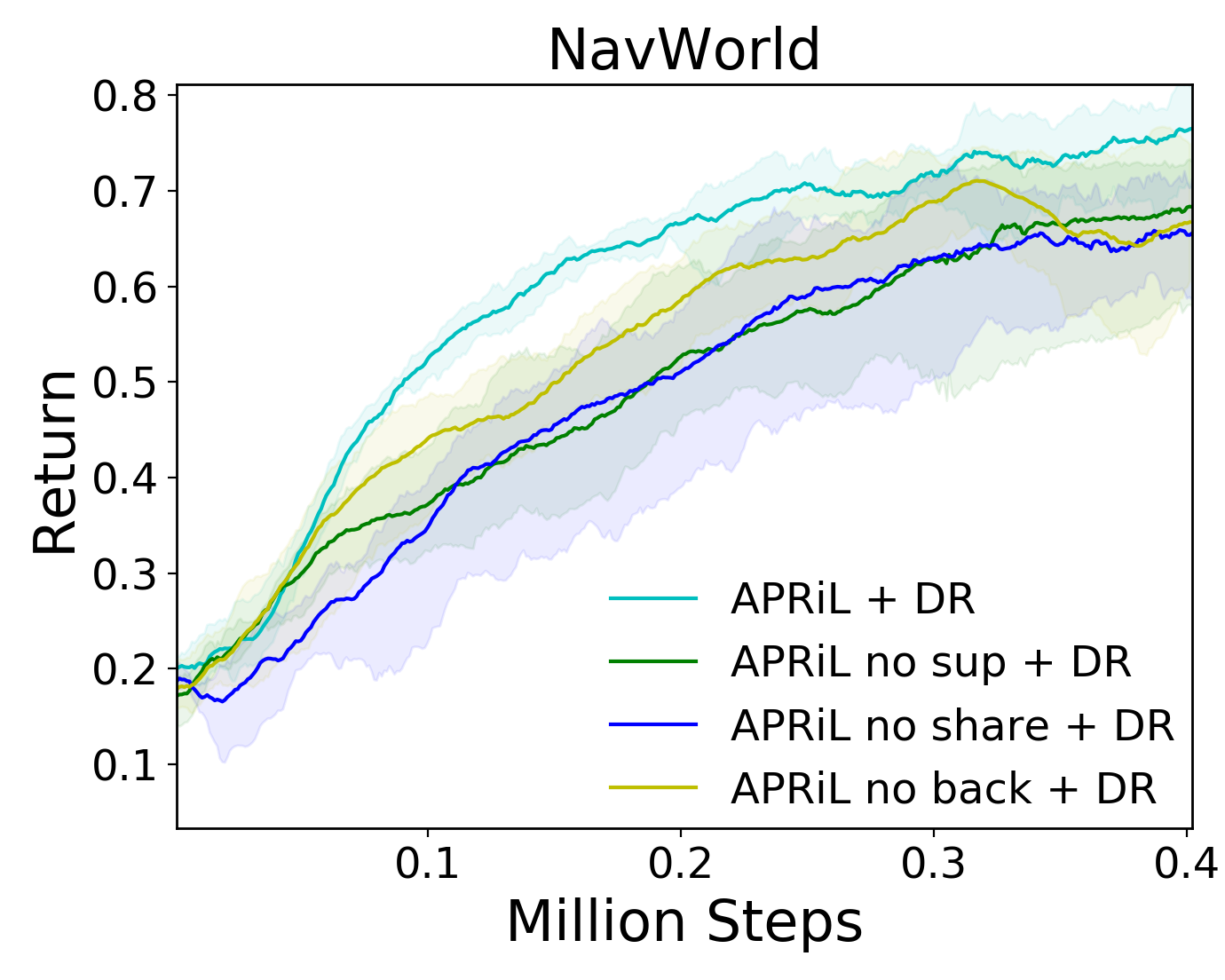} 
    \includegraphics[width=0.32\linewidth]{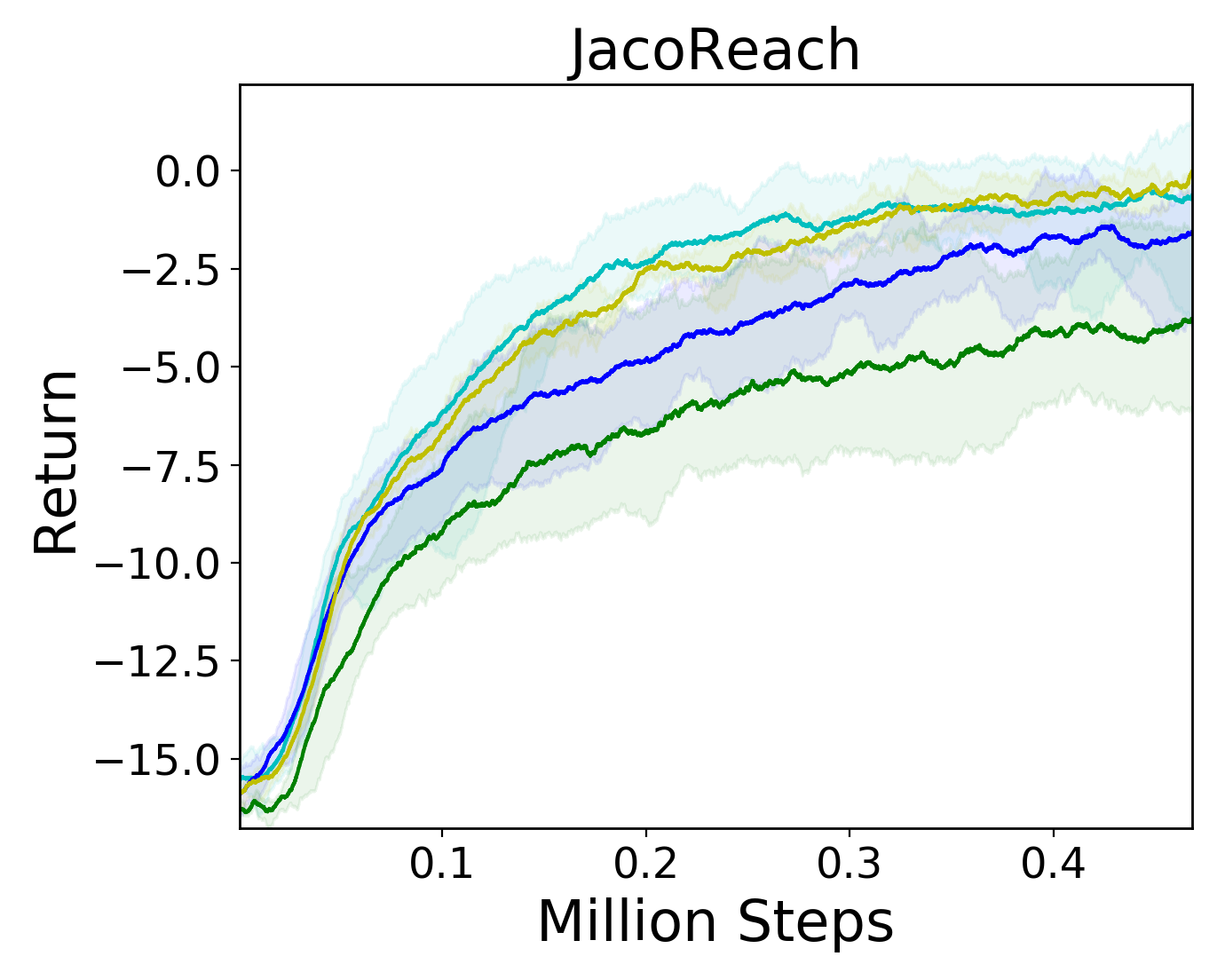}
    \includegraphics[width=0.32\linewidth]{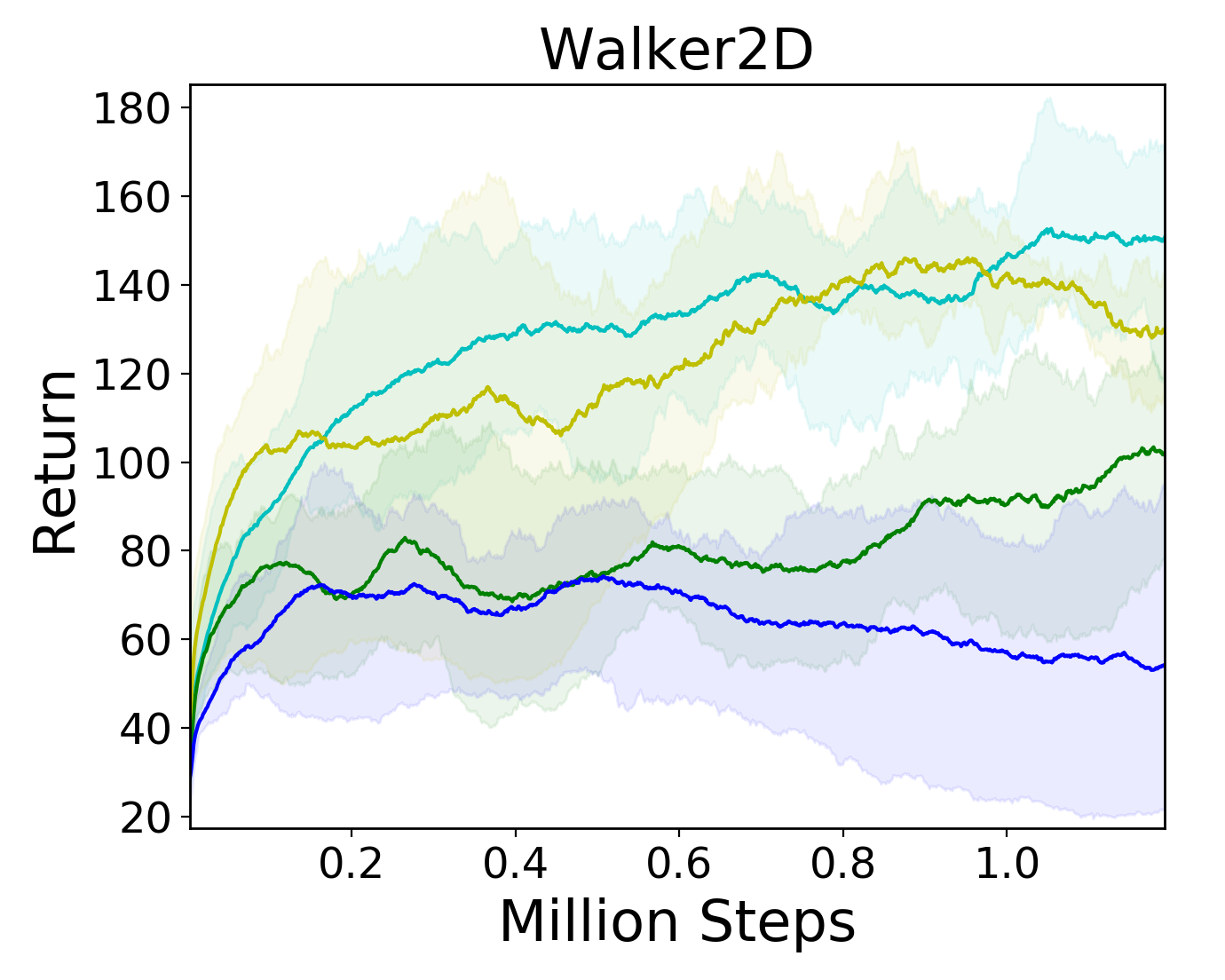}

    \caption{Learning curves for observation-based policies during Domain Randomisation (DR). \textbf{Top row}: comparison with baselines. \textbf{Bottom row}: comparison with ablations. \textbf{Solid line}: mean performance. \textbf{Shaded region}: covers minimum and maximum performances across $5$ seeds. \met's attention and shared replay lead to stronger or commensurate performance.}
    \label{fig:SARL_experiments}
\end{figure*}

\begin{figure*}
    \centering
    \includegraphics[width=0.36\linewidth]{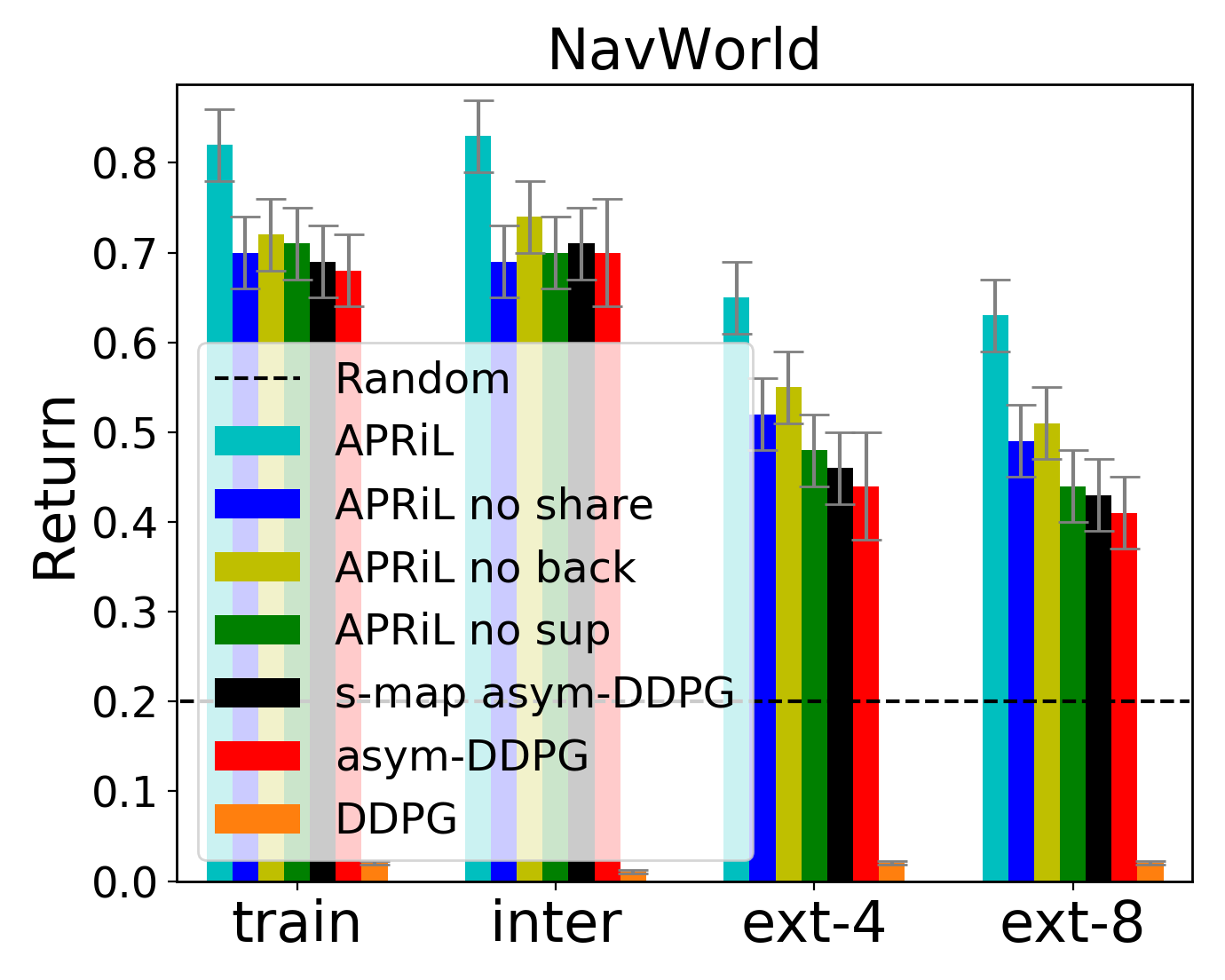}
    \includegraphics[width=0.36\linewidth]{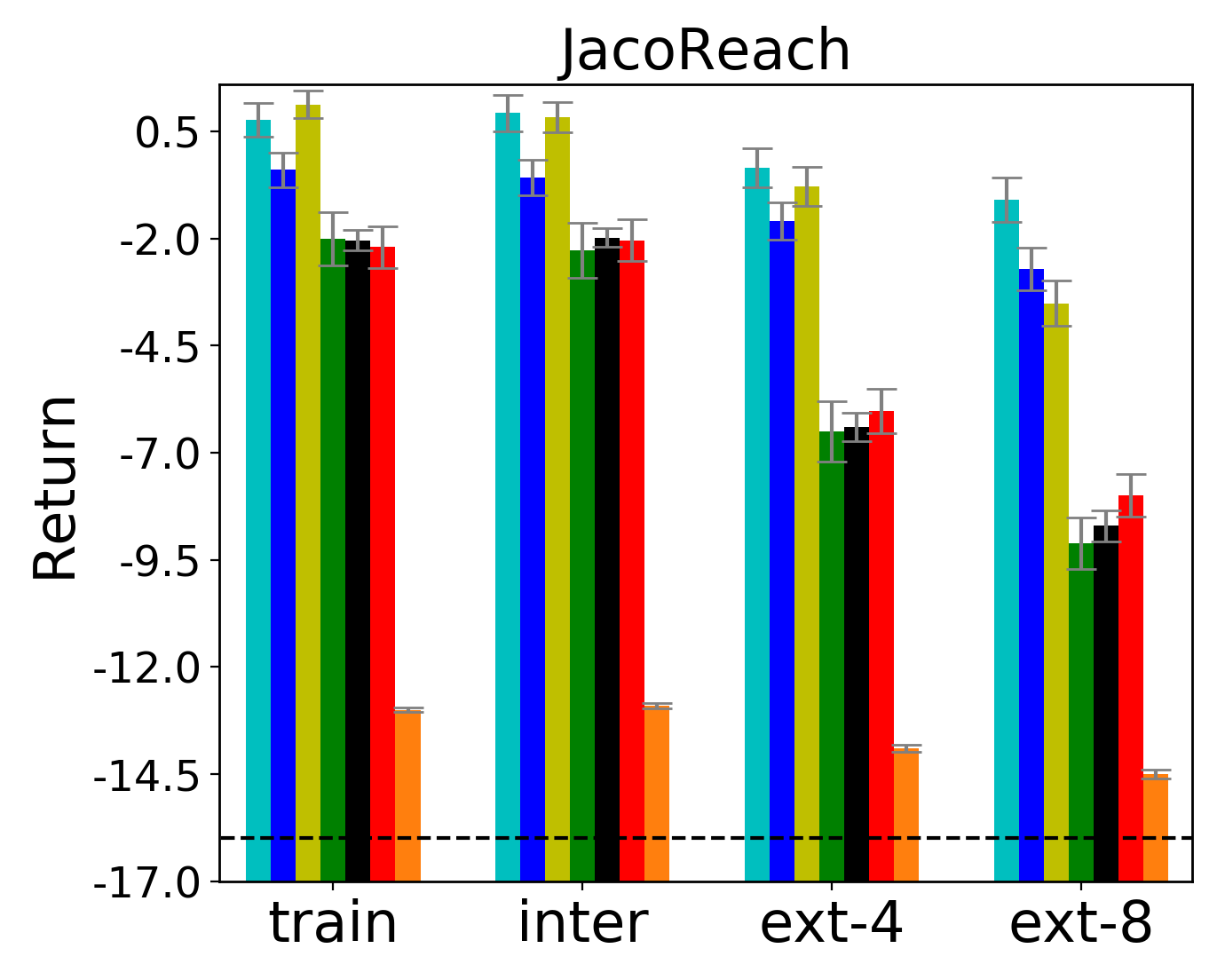}
    \includegraphics[width=0.18\linewidth]{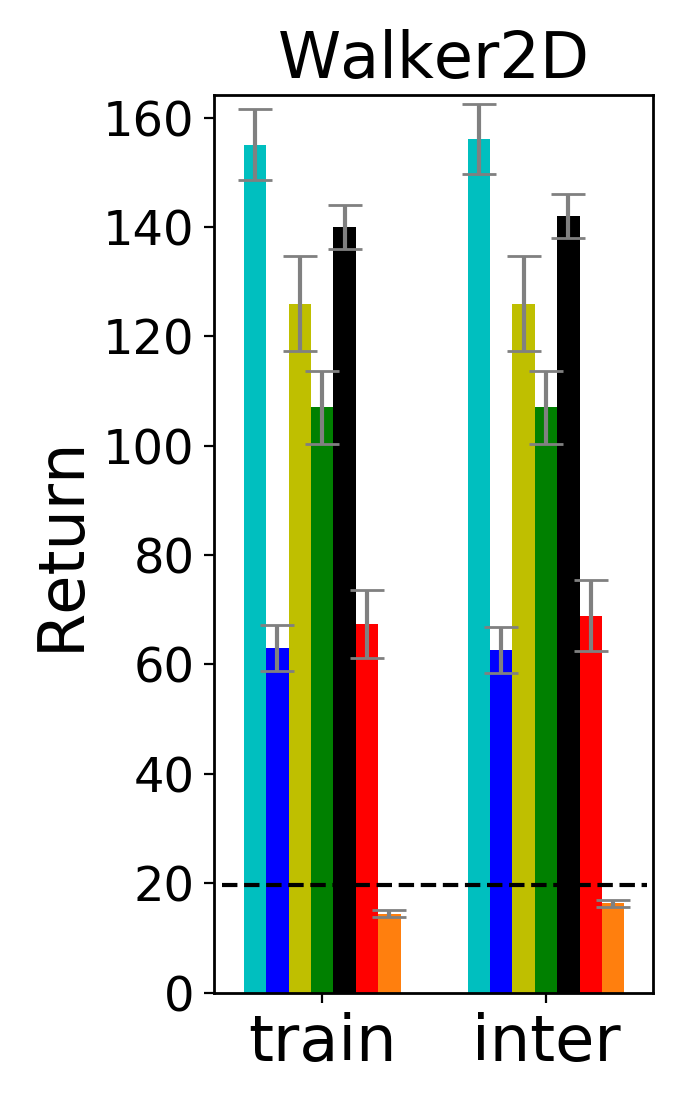}
    \caption{Comparing average return of the image-policy between training, interpolated and extrapolated domains (100 each). Plots reflect mean and 2 standard deviations for average return (5 seeds). \met~generalises due to its attention and outperforms the baselines. We compare against a random agent to gauge the degree of degradation in policy performance between domains.}
    \label{fig:percent_decrease}
\end{figure*}
\begin{figure*}
    \centering
    \includegraphics[width=0.99\linewidth]{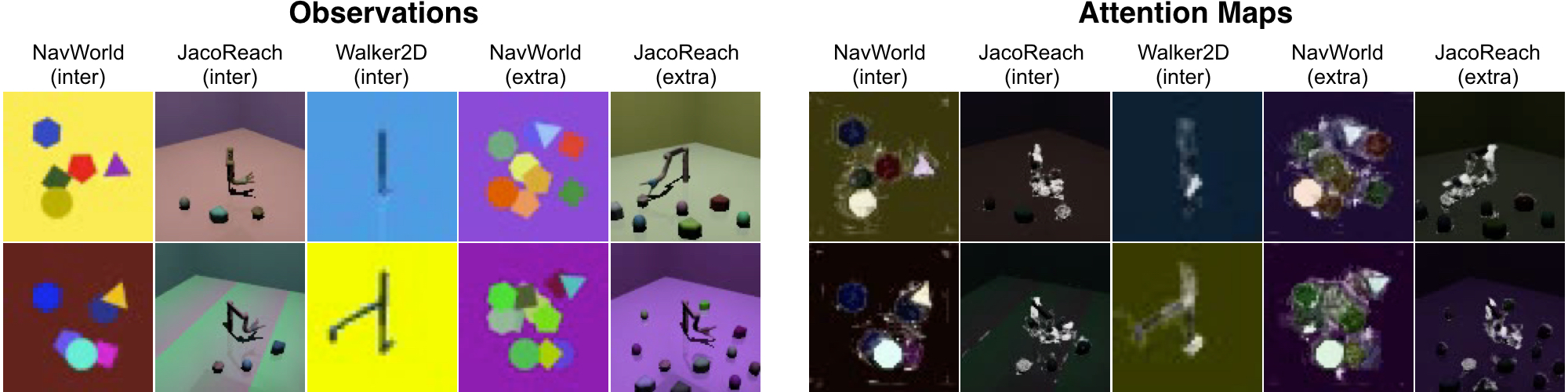}
    \caption{Held-out domains and \met~attention maps. For the extrapolated domain columns (extra), top and bottom represent \textbf{ext-4} and \textbf{ext-8}. White/black signify high/low attention values. Attention suppresses the background and distractors and helps generalise.}
    \label{fig:hold_out_envs}
\end{figure*}

\label{sec:protocol}

We evaluate \met~over the following environments (see Appendix~\ref{app:environments} for more details): {1. \textit{NavWorld}: the circular agent is sparsely rewarded for reaching the triangular target in the presence of distractors.} {2. \textit{JacoReach}: the Kinova arm is rewarded for reaching the diamond-shaped object in the presence of distractors.} {3. \textit{Walker2D}: this slightly modified (see Appendix~\ref{app:environments}) Deepmind Control Suite environment~\citep{tassa2018deepmind} the agent is rewarded for walking forward whilst keeping its torso upright.}

During training, for \met~, its ablations, and all baselines, we perform Domain Randomisation (DR) \cite{tobin2017domain, sadeghi2016cad2rl}, randomising the following environment parameters to enable generalisation with respect to them: camera position, orientation, textures, materials, colours, object locations, background (see Appendix~\ref{app:rand}).

We start by comparing \met~against two competitive baselines that also exploit privileged information during training. We compare against the \textit{Asymmetric DDPG} (asym-DDPG) baseline \citep{pinto2017asymmetric} to evaluate the importance of privileged attention and shared replay for learning and robustness to distractors. Our second baseline, \textit{State-Mapping Asymmetric DDPG} (s-map asym-DDPG), introduces a bottleneck layer trained to predict the environment state using an \textit{$L_{2}$} loss. This is another intuitive approach that further exploits state information in simulation~\citep{zhang2016vision} to learn informative representations that are robust to visual randomisation. This approach does not incorporate object-centric attention or leverage privileged object segmentations. We note that since this baseline learns state estimation it is not expected to extrapolate well to domains with additional distractor objects and varying state spaces (with respect to the training domain). We also compare \met~with DDPG to emphasise the difficulty of these DR tasks if privilege information is not leveraged. 

We perform an ablation study to investigate which components of \met~contribute to performance gains. The ablations consist of: {1. \textit{\met~no sup}: the full setup except without attention alignment. Here the observation attention module must learn without guidance from the state-agent.} {2. \textit{\met~no share}: \met~without a shared replay.} {3. \textit{\met~no back}: uniform object attention values $c$ are used to train the observation attention module, thereby only suppressing the background. Here we investigate the importance of object-suppression for generalisation.}

We investigate the following to evaluate how well \met~facilitates transfer across visually distinct domains: Does~\met~:
{1. Increase \textbf{sample-efficiency} during training?} 
{2. Affect \textbf{interpolation} performance on unseen environments from the training distribution?}
{3. Affect \textbf{extrapolation} performance on environments outside the training distribution?}

\subsection{Performance On The Training Distribution}

Figure \ref{fig:SARL_experiments} shows that \met~outperforms the baselines for each environment (except \textit{Walker2D} where it matches s-map asym-DDPG). The ablations in Figure \ref{fig:SARL_experiments} show that a shared replay buffer, background suppression, and attention-alignment each individually provide benefits but are most effective when combined together. Interestingly, background suppression is extremely effective for sample-efficiency, as for these domains the majority of the irrelevant, highly varying, aspects of the observation-space are occupied by the background. It is also surprising that the s-map asym-DDPG baseline, which learns to map to environment states, does not outperform asym-DDPG and does not match \met~'s
performance for \textit{NavWorld} and \textit{JacoReach}. For these domains, predicting states (including those of distractors) is difficult\footnote{For \textit{JacoReach} prediction errors and policy performance are sensitive to state-space. In Figure \ref{fig:SARL_experiments} we plot the best performing state-space. Refer to Appendix \ref{app:s-map-abl} for further details.} and prediction errors limit policy performance. For \textit{Walker2D}, in the absence of distractor objects, s-map asym-DDPG is a competitive baseline and \met~provides
marginal gains.

\subsection{Interpolation: Transfer To Domains From The Training Distribution}
\label{interp}

We evaluate performance on environments unseen during training but within the training distribution (see Appendix \ref{app:rand}). For \textit{NavWorld} and \textit{JacoReach}, the interpolated environments have the same number of distractors, sampled from the same object catalogue, as the training distribution. Figure \ref{fig:percent_decrease} plots the return on these held-out domains. For all algorithms, we observe minimal degradation in performance between training and interpolated domains. However, as \met~outperforms on the training distribution (apart from; \textit{Walker2D} for s-map asym-DDPG, \textit{JacoReach} for \met~no back), its final performance on the interpolated domains is significantly better, emphasising the benefits of both privileged attention and a shared replay.

\subsection{Extrapolation: Transfer To Domains Outside The Training Distribution}
\label{extrap}

For \textit{NavWorld} and \textit{JacoReach}, we investigate how well each method generalises to extrapolated domains with additional distractor objects (specifically 4 or 8; referred as \textbf{ext-4} and \textbf{ext-8}). The textures and colours of these objects are sampled from a held-old out set not seen during training. The locations are sampled; randomly for \textit{NavWorld}, from extrapolated arcs of two concentric circles of different radii for \textit{JacoReach}. Shapes are sampled from the training catalogue of distractors. We do not extrapolate for \textit{Walker2D}, as this domain does not contain distractors. However, we show (in the previous sections) that \met~is still beneficial during DR for this domain and therefore demonstrate its application does not need to be restricted to environments with clutter. Please refer to Figure \ref{fig:hold_out_envs} for examples of the extrapolated domains.

Figure \ref{fig:percent_decrease} shows that \met~generalises and performs considerably better on the held-out domains than each baseline. Specifically, when comparing with the baselines that leverage privilege information, for \textit{JacoReach} performance falls by $11$\%\footnote{Percentage decrease is taken with respect to additional return over a random agent
on the training domain.} for \met~instead of $42$\% and $48$\% for asym-DDPG and s-map asym-DDPG respectively. The ablations demonstrate that effective distractor suppression is crucial for generalisation. This is particularly prominent for \textit{JacoReach} where the performance drop for the methods that use attention-alignment (\met~and \met~no share) is $11$\% and $15$\%, which is far less than $27$\% and $51$\% (\met~no back and \met~no sup) for those that do not learn to effectively suppress distractors.

\subsection{Attention Module Analysis}
\label{attention_analysis}

We visualise \met's attention maps (Figure \ref{fig:hold_out_envs}, \ref{fig:ATTMAPS_2}, \ref{fig:ATTMAPS} (in Appendix \ref{app:s-map-abl}) and these \href{https://sites.google.com/view/april-domain-randomisation/home}{videos}) on both interpolated and extrapolated domains. For \textit{NavWorld}, attention is correctly paid to all relevant aspects (agent and target; circle and triangle respectively) and generalises well. For \textit{JacoReach}, attention suppresses the distractors even on the extrapolated domains, achieving robustness with respect to them. Interestingly, as we encourage sparse attention, \met~learns to only pay attention to every-other-link of the arm (as the state of an unobserved link can be inferred by observing those of the adjacent links). For \textit{Walker2D}, dynamic object attention is learnt (different objects are attended based on the state of the system - see Figure \ref{fig:ATTMAPS}). When upright, walking, and collapsing, \met~pays attention to the lower limbs, every other link, and foot and upper body, respectively. We suspect that in these scenarios, the optimal action depends most on the state of the lower links (due to stability), every link (coordination), and foot and upper body (large torque required), respectively.

\section{Related Work}
A large body of work investigates the problem of learning robust policies that generalise well outside of the training distribution.
Work on \textbf{transfer learning} leverages representations from one domain to efficiently solve a problem from a different domain~\citep{donahue2014decaf, oquab2014learning, rusu2016sim}.
In particular, \textbf{domain adaptation} techniques aim to adapt a learned model to a specific target domain, often optimising models such that representations are invariant to the shift in the target domain ~\citep{ganin2016domain, long2015learning, bousmalis2016domain, wulfmeier2017mutual}.
These methods commonly require data from the target domain in order to transfer and adapt effectively.

In contrast, \textbf{domain randomisation} covers a distribution of environments by randomising visual~\citep{tobin2017domain} or dynamical parameters~\citep{peng2018sim} during training in order to generalise~\citep{sadeghi2016cad2rl, rajeswaran2016epopt, viereck2017learning,held2017probabilistically, openai2018handmanip, laskin2020reinforcement}.
In doing so, such methods shift the focus from adaptation to specific environments to \textbf{generalisation and robustness} by covering a wide range of variations.
Recent work automatically varies this distribution during training~\citep{akkaya2019solving} or trains a canonical invariant image representation~\citep{james2019sim}.
However, while randomisation can enable us to learn robust policies, it significantly increases training time due to the increased environment variability~\citep{openai2018handmanip}, and can reduce asymptotic performance. Our work partially addresses this fact by training two agents, one of which is not affected by visual randomisations.

Other works explicitly encourage representations \textbf{invariant} to observation space variations \citep{slaoui2019robust, srinivas2020curl, james2019sim}. Contrastive techniques \citep{slaoui2019robust, srinivas2020curl} use a clear separation between positive (and negative) examples, predefined by the engineer, to encourage \textbf{invariance}. Unlike \met, these \textbf{invariances} are over abstract spaces and are not designed to exploit \textbf{privileged information}; shown to be beneficial by \met's ablations. Furthermore, \met's \textbf{invariance} is \textbf{task-driven} via attention. Approaches like \citep{james2019sim,zhang2016vision}, learn invariances in a supervised manner, mapping from observation to a predefined space. Unlike \met, these methods are unable to discover task-independent aspects of the mapping-space, limiting robustness and generalisation. Finally, unlike \met~as a model-free RL approach, some model-based works use forward or inverse models \citep{pathak2017curiosity, hafner2019learning, schmeckpeper2019learning} to achieve invariance. 

Existing comparisons in the literature demonstrate that, even without domain randomisation, the increased dimensionality and potential partial observability complicates learning for RL agents~\citep{tassa2018deepmind,schwab2019simultaneously}. In this context, accelerated training has also been achieved by using \textbf{access to privileged information} such as environment states to asymmetrically train the critic in actor-critic RL~\citep{schwab2019simultaneously,pinto2017asymmetric,foerster2018counterfactual}.
In addition to using additional information to train the critic, \citep{schwab2019simultaneously} use a \textbf{shared replay buffer} for data generated by image- and state-based actors to further accelerate training for the image-based agent. 
Our method extends these approaches by sharing information about relevant objects by aligning agent-integrated attention mechanisms between an image- and state-based actors.

Recent experiments have demonstrated the strong dependency and interaction between attention and learning in human subjects~\citep{LEONG2017451}. In the context of machine learning, \textbf{attention mechanisms} have been integrated into RL agents to increase robustness and enable interpretability of an agent's behaviour~\citep{sorokin2015deep, Mott2019TowardsIR}. In comparison, we focus on utilising the attention mechanism as an interface to transfer information between two agents to enable faster training and better generalisation.

\section{Conclusion}
\label{conclusion}

We introduce \method~(\met), an extension to asymmetric actor-critic algorithms that leverages attention mechanisms and access to privileged information such as simulator environment states. The method benefits in two ways in addition to asymmetry between actor and critic: via aligning attention masks between image- and state-space agents, and by sharing a replay buffer. 
Since environment states are not affected by visual randomisation, we are able to learn efficiently in the image domain especially during domain randomisation where feature learning becomes increasingly difficult. 
Evaluation on a diverse set of environments demonstrates significant improvements over competitive baselines including asym-DDPG and s-map asym-DDPG; and show that \met~learns to generalise favourably to environments not seen during training (both within and outside of the training distribution). Finally, we investigate the relative importance of the different components of \met~ in an extensive ablation.

\bibliography{example}

\clearpage
\appendix
\section{Environments}
\label{app:environments}

\begin{enumerate}
\item \textit{NavWorld}: In this sparse reward, 2D environment, the goal is for the circular agent to reach the triangular target in the presence of distractor objects. Distractor objects have 4 or more sides and apart from changing the visual appearance of the environment cannot affect the agent. The state space consists of the $[x,y]$ locations of all objects. The observation space comprises RGB images of dimension $(60 \times 60 \times 3)$. The action space corresponds to the velocity of the agent. The agent only obtains a sparse reward of $+1$ if the particle is within $\epsilon$ of the target, after which the episode is terminated prematurely. The maximum episodic length is 20 steps, and all object locations are randomised between episodes.

\item \textit{JacoReach}: In this 3D environment the goal of the agent is to move the Kinova arm such that the distance between its hand and the diamond-shaped object is minimised. The state space consists of the quaternion position and velocity of each joint as well as the Cartesian positions of each object. The observation space comprises RGB images and is of dimension $(100 \times 100 \times 3)$. The action space consists of the desired relative quaternion positions of each joint (excluding the digits) with respect to their current positions. Mujoco uses a PD controller to execute 20 steps that minimises the error between each joint's actual and target positions. The agent's reward is the negative squared Euclidean distance between the Kinova hand and diamond object plus an additional discrete reward of $+5$ if it is within $\epsilon$ of the target. The episode is terminated early if the target is reached. All objects are out of reach of the arm and equally far from its base. Between episodes the locations of the objects are randomised along an arc of fixed radius with respect to the base of the Kinova arm. The maximum episodic length is 20 agent steps.

\item \textit{Walker2D}: In this 2D modified Deepmind Control Suite environment~\citep{tassa2018deepmind} with a continuous action-space the goal of the agent is to walk forward as far as possible within $300$ steps. We introduce a limit to episodic length as we found that in practice this helped stabilise learning across all tested algorithms. The observation space comprises of $2$ stacked RGB images and is of dimension $(40 \times 40 \times 6)$. Images are stacked so that velocity of the walker can be inferred. The state space consists of quaternion position and velocities of all joints. The absolute positions of the walker along the x-axis is omitted such that the walker learns to become invariant to this. The action space is setup in the same way as for the \textit{JacoReach} environment. The reward is the same as defined in~\citep{tassa2018deepmind} and consists of two multiplicative terms: one encouraging moving forward beyond a given speed, the other encouraging the torso of the walker to remain as upright as possible. The episode is terminated early if the walker's torso falls beyond either $[-1,1]$ radians with the vertex or $[0.8,2.0]$m along the z axis. 
\end{enumerate}

\section{Randomisation Procedure}
\label{app:rand}

In this section we outline the randomisation procedure taken for each environment during training.
\begin{enumerate}
\item \textit{NavWorld}: Randomisation occurs at the start of every episode. We randomise the location, orientation and colour of every object as well as the colour of the background. We therefore hope that our agent can become invariant to these aspects of the environment.

\item \textit{JacoReach}: Randomisation occurs at the start of every episode. We randomise the textures and materials of every object, Kinova arm and background. We randomise the locations of each object along an arc of fixed radius with respect to the base of the Kinova arm. Materials vary in reflectance, specularity, shininess and repeated textures. Textures vary between the following: noisy (where RGB noise of a given colour is superimposed on top of another base colour), gradient (where the colour varies linearly between two predefined colours), uniform (only one colour). Camera location and orientation are also randomised. The camera is randomised along a spherical sector of a sphere of varying radius whilst always facing the Kinova arm. We hope that our agent can become invariant to these randomised aspects of the environment.

\item \textit{Walker2D}: Randomisation occurs at the start of every episode as well as after every $50$ agent steps. We introduce additional randomisation between episodes due to their increased duration. Due to the MDP setup, intra-episodic randomisation is not an issue. Materials, textures, camera location and orientation, are randomised in the same procedure as for \textit{JacoReach}. The camera is setup to always face the upper torso of the walker.
\end{enumerate}

\section{Implementation details}
\label{hyperparams}

In this section we provide more details on our training setup. Refer to table \ref{table:parameters} for the model architecture for each component of \met~and the asymmetric DDPG baseline. \textit{Obs Actor} and \textit{Obs Critic} setup are the same for both \met~and the asymmetric DDPG baseline. \textit{Obs Actor} model structure comprises of the convolutional layers (without padding) defined in  table \ref{table:parameters} followed by one fully connected layer with $256$ hidden units (FC($\lbrack256\rbrack$)). The state-mapping asymmetric DDPG baseline has almost the same architecure as \textit{Obs Actor}, except there is one additional fully connected layer, directly after the convolutional layers that has the same dimensions as the environment state space. When training this intermediate layer on the $L_2$ state regressor loss, the state targets are normalised using a running mean and standard deviation, similar to DDPG, to ensure each dimension is evenly weighted and to stabilise targets. The DDPG baseline has the same policy architecture as the other baselines except now the critic is image-based and has the same structure as the actor. All layers use ReLU activations and layer normalisation unless otherwise stated. Each actor network is followed by a tanh activation and rescaled to match the limits of the environment's action space.

\begin{table}[h!]
\caption{Model architecture. FC() and Conv() represent a fully connected and convolutional network. The arguments of FC() and Conv() take the form [nodes] and [channels, square kernel size, stride] for each hidden layer respectively.}
% \scriptsize
\begin{center}
\begin{tabular}{ |c|c|c| } 
\hline
 Domain & NavWorld and JacoReach & Walker2D\\
\hline
State Actor & FC($\lbrack256\rbrack$) & FC($\lbrack256\rbrack$)\\ 
Obs Actor & Conv($\lbrack\lbrack18,7,1\rbrack,\lbrack32,5,1],$ &  Conv($\lbrack\lbrack18,8,2\rbrack,\lbrack32,5,1\rbrack,$ \\ 
 &$\lbrack32,3,1\rbrack\rbrack$)& $\lbrack16,3,1\rbrack,\lbrack4,3,1\rbrack\rbrack$)\\
\hline
State Critic  & FC($\lbrack64, 64\rbrack$) & FC($\lbrack400, 300\rbrack$)\\ 
Obs Critic & FC($\lbrack64, 64\rbrack$) & FC($\lbrack400, 300\rbrack$)\\ 
\hline
State Attention & FC($\lbrack256\rbrack$) & FC($\lbrack256\rbrack$)\\ 
Obs Attention & Conv($\lbrack\lbrack32,8,1\rbrack,\lbrack32,5,1],$ & Conv($\lbrack\lbrack32,8,1\rbrack,\lbrack32,5,1],$\\ 
&$\lbrack64,3,1\rbrack\rbrack$)&$\lbrack64,3,1\rbrack\rbrack$)\\
\hline
Replay Size & $10^4$ & $2 \times 10^5$\\ 
\hline
\end{tabular}
\label{table:parameters}
\end{center}
\end{table}

The \textit{State Attention} module includes the fully connected layer defined in table \ref{table:parameters} followed by a Softmax operation. The \textit{Obs Attention} module has the convolutional layers (with padding to ensure constant dimensionality) outlined in table \ref{table:parameters} followed by a fully connected convolutional layer (Conv($\lbrack1,1,1\rbrack$)) with a Sigmoid activation to ensure the outputs vary between $0$ and $1$. The output of this module is tiled in order to match the dimensionality of the observation space.

During each iteration of \met~(for both $\omodule$ and $\smodule$) we perform $50$ optimization steps on minibatches of size $64$ from the replay buffer. The target actor and critic networks are updated with a Polyak averaging of $0.999$. We use Adam optimizer with learning rate of $10^{-3}$, $10^{-4}$ and $10^{-4}$ for critic, actor and attention networks. We use default TensorFlow values for the other hyperparameters. The discount factor, entropy weighting and self-supervised learning hyperparameters are $\gamma = 0.99$, $\beta = 0.0008$ and $\nu = 1$. To stabilize learning, all input states are normalized by running averages of the means and standard deviations of encountered states. Both actors employ adaptive parameter noise~\citep{plappert2017parameter} exploration strategy with initial std of $0.1$, desired action std of $0.1$ and adoption coefficient of $1.01$. The settings for the baseline are kept the same as for \met~where appropriate.

\section{Attention Visualisation}

\begin{figure*}[ht!]
    \centering
    \includegraphics[width=0.9\linewidth]{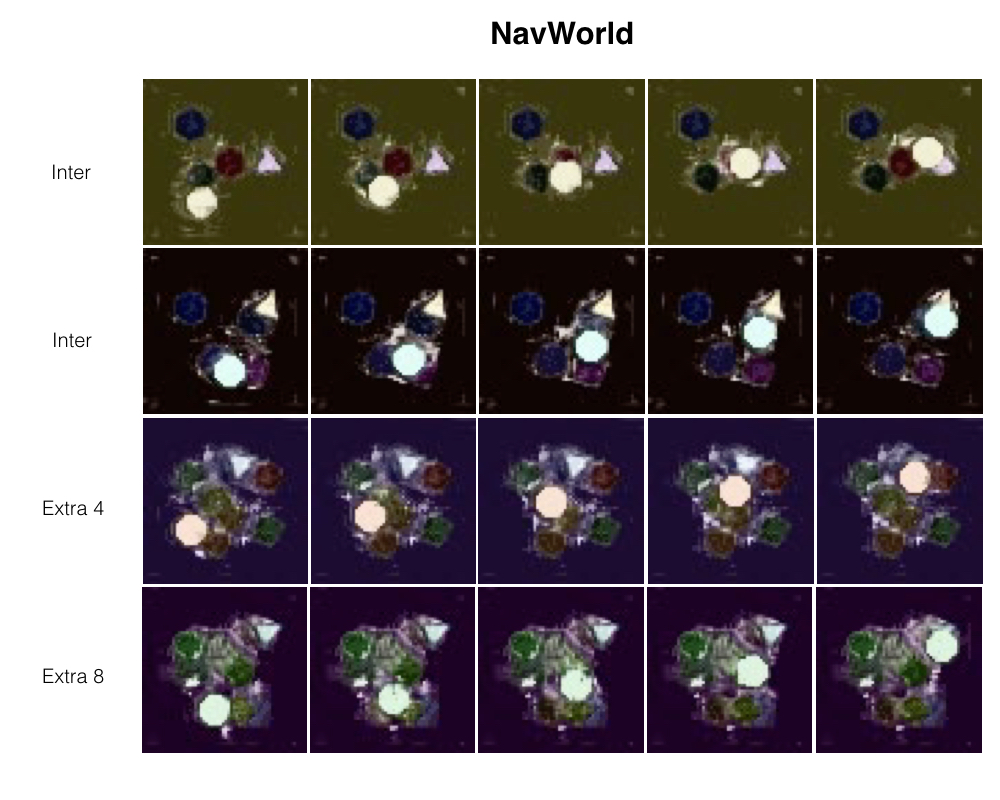}
    \includegraphics[width=0.9\linewidth]{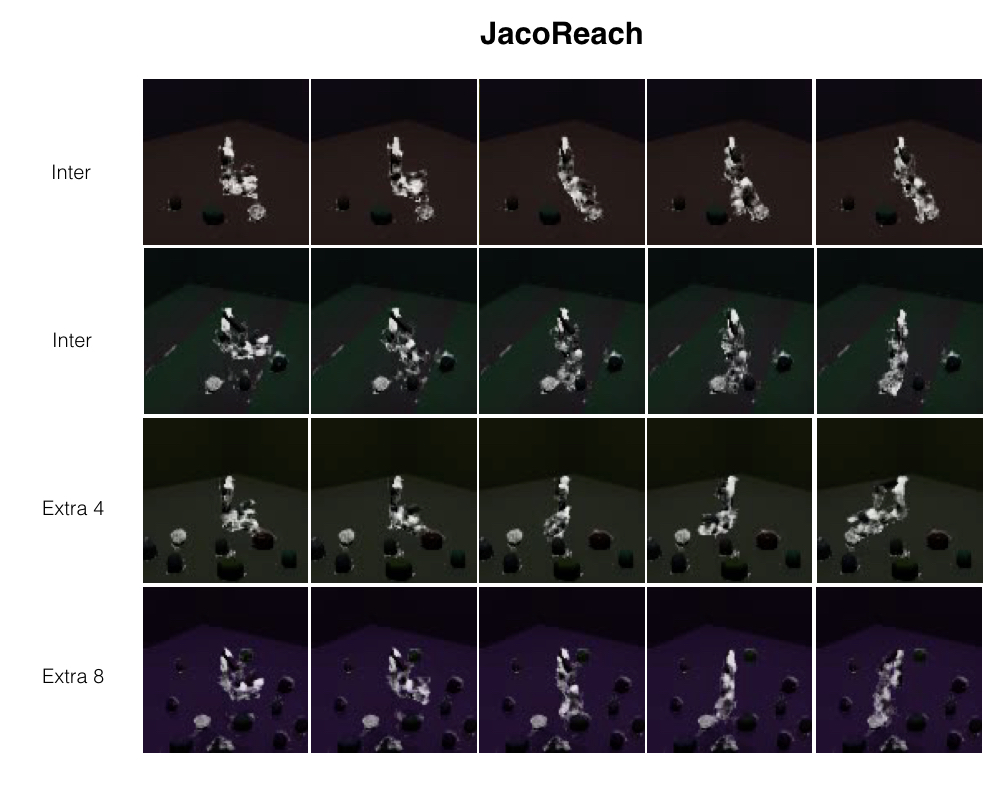}
    \caption{\met~attention maps for policy rollouts on NavWorld and Jaco domains. White and black signify high and low attention values respectively. For NavWorld and JacoReach, attention is correctly paid only to the relevant objects (and Jaco links), even for the extrapolated domains. Refer to section \ref{attention_analysis} for more details.}
    \label{fig:ATTMAPS_2}
\end{figure*}

\begin{figure*}[ht!]
    \centering
    \includegraphics[width=0.9\linewidth]{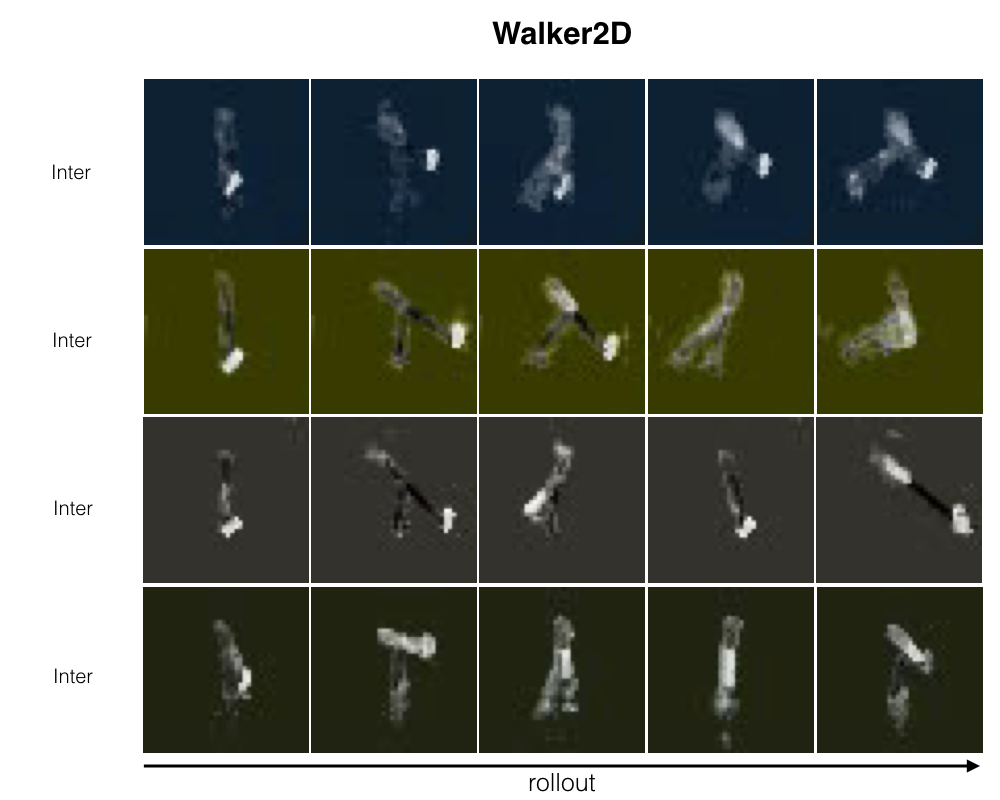}
    \caption{\met~attention maps for policy rollouts on Walker domain. White and black signify high and low attention values respectively. Attention varies based on the state of the walker. When the walker is upright, high attention is paid to lower limbs. When walking, even attention is paid to every other limb. When about to collapse, high attention is paid to the foot and upper torso. Refer to section \ref{attention_analysis} for more details.}
    \label{fig:ATTMAPS}
\end{figure*}

Figures (\ref{fig:ATTMAPS_2}, \ref{fig:ATTMAPS}) show \met's attention maps for policy roll-outs on each environment and held-out domain. Attention attends to the task-relevant objects and generalises well. 

\section{State Mapping Asymmetric DDPG Ablation Study}
\label{app:s-map-abl}
\begin{wrapfigure}{r}{0.5\textwidth}
    \centering
       \includegraphics[width=0.90\linewidth]{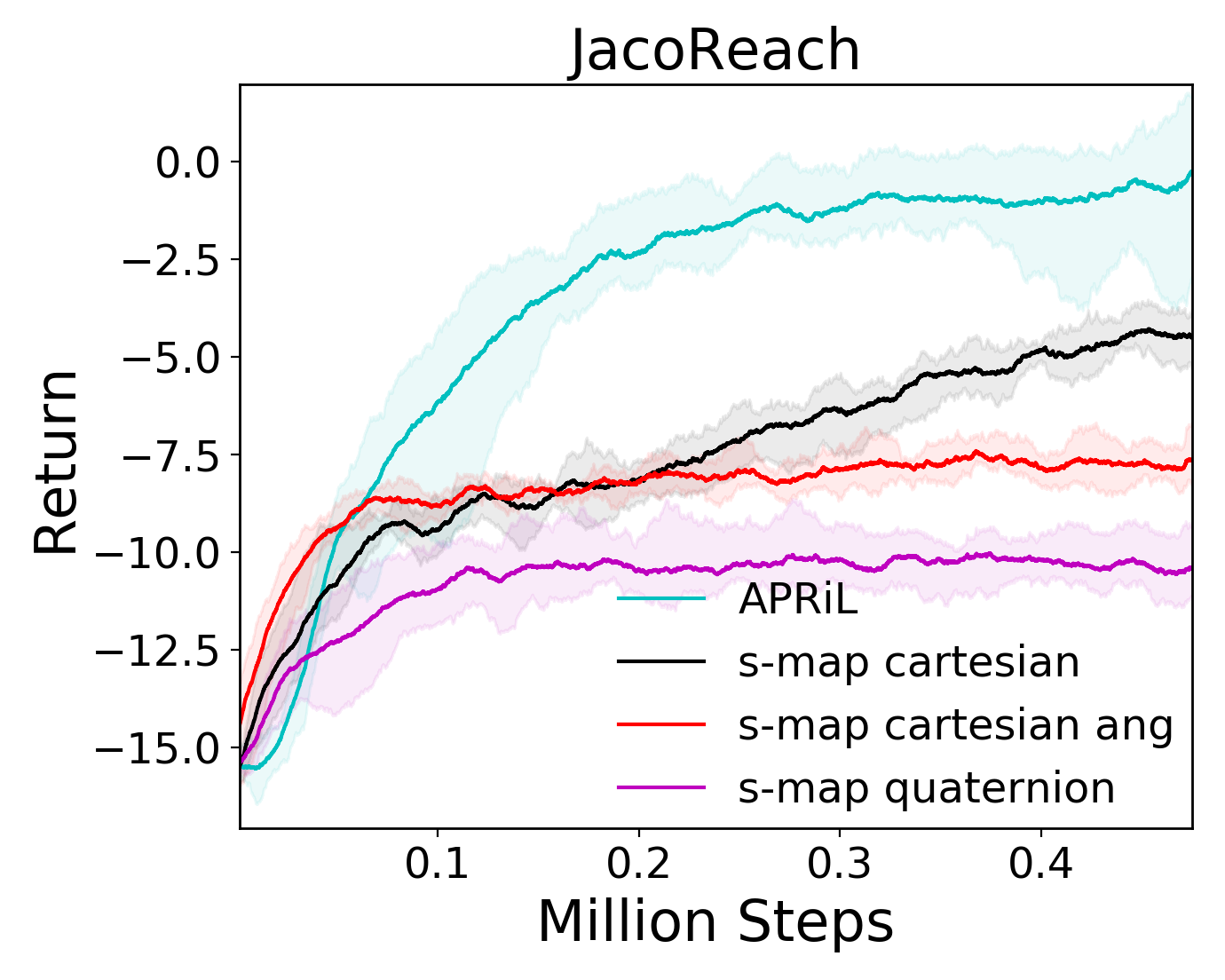}
    \caption{We compare learning of \met~with variants of s-map asym-DDPG. For \textbf{s-map cartesian}, \textbf{s-map cartesian ang} and \textbf{s-map quaternion}, regressed states are cartesian position, cartesian position and rotation, and quaternions respectively (for Jaco arm - distractors are always cartesian).}
    \label{fig:s-map_ablation}
\end{wrapfigure}
\begin{figure*}[ht!]
    \centering
     \includegraphics[width=0.252\linewidth]{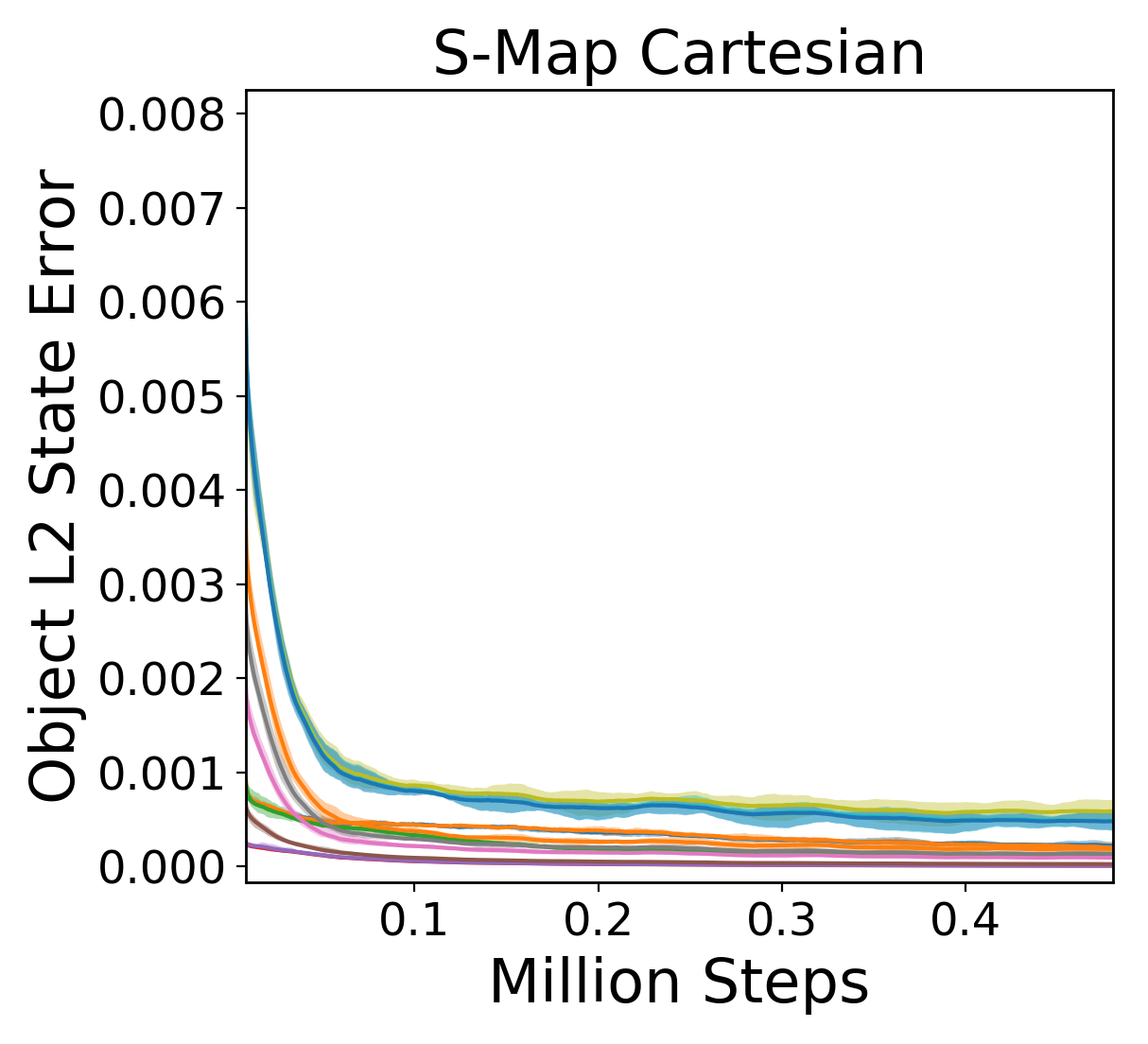}
      \includegraphics[width=0.252\linewidth]{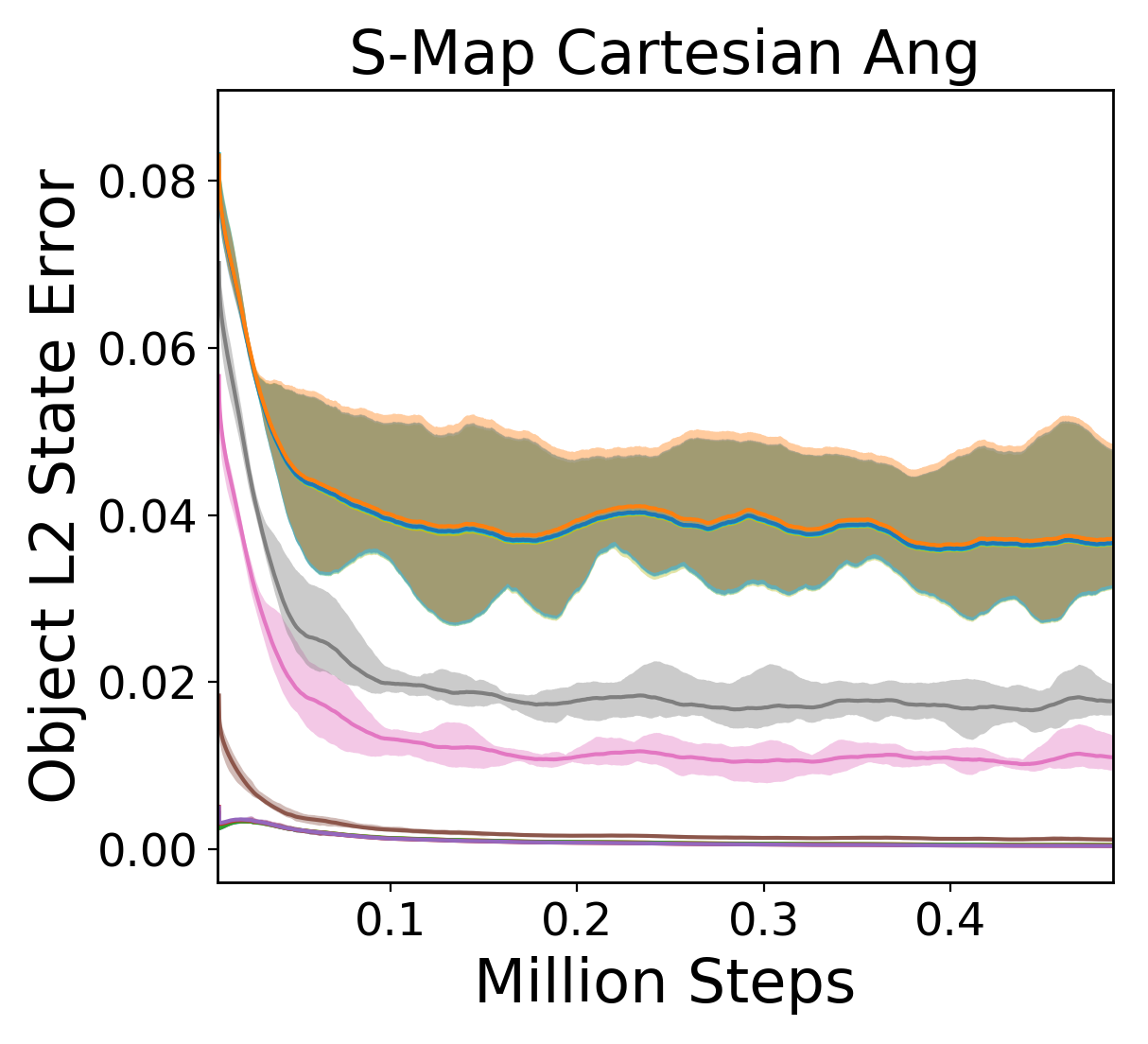}
      \includegraphics[width=0.457\linewidth]{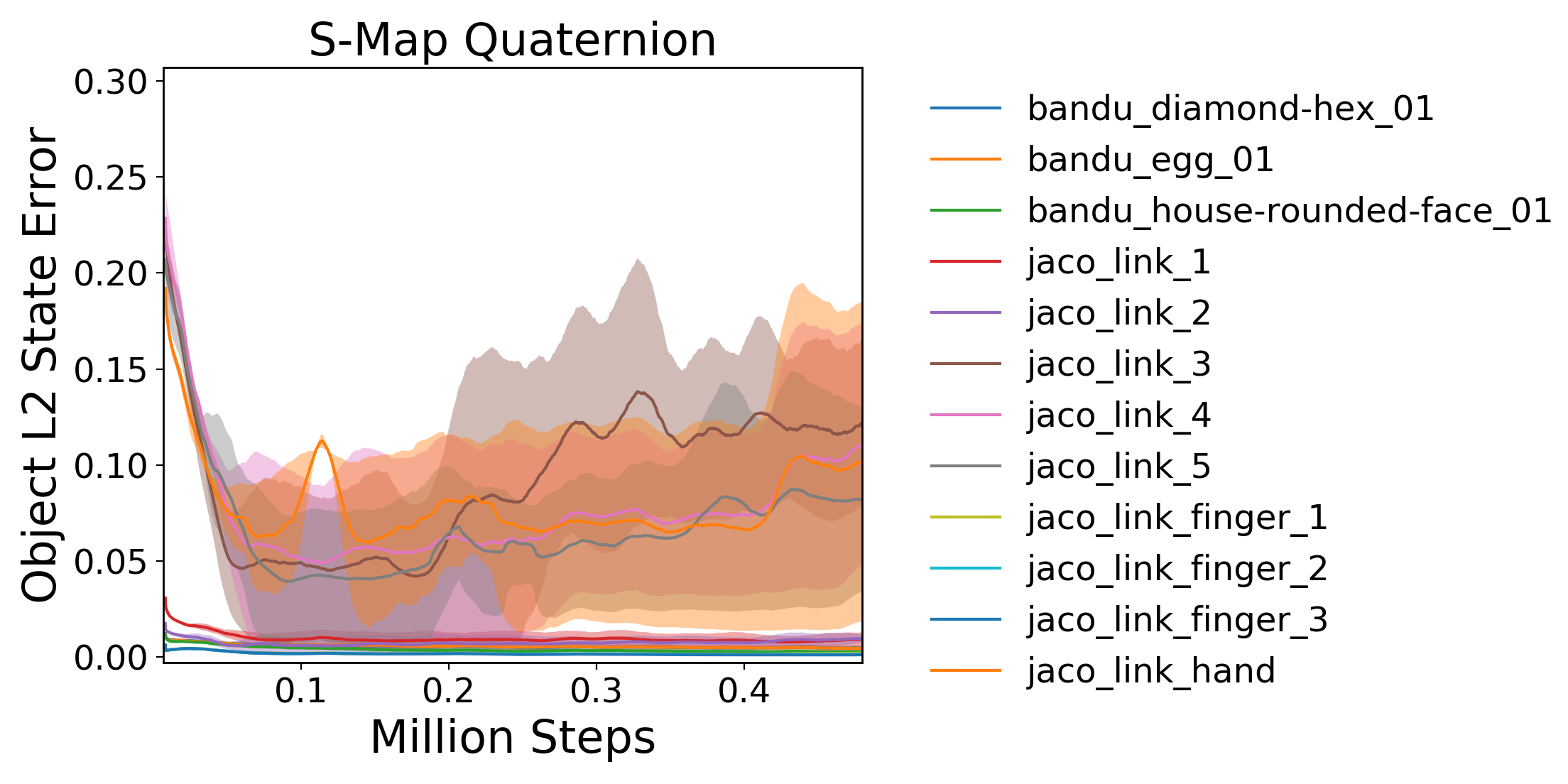}
    \caption{S-Map Asym-DDPG normalised state prediction errors. We compare individual object $L_2$ regressor losses (mean loss over states corresponding to a given object) between \textbf{s-map cartesian}, \textbf{s-map cartesian ang} and \textbf{s-map quaternion}. The object keys are on the right. \textbf{S-map quaternion} and \textbf{s-map cartesian ang} struggle to regress to quaternions and cartesian rotations and hence policy performance is restricted. }
    \label{fig:hold_out_envs_rotation}
\end{figure*}

We found that for \textit{JacoReach}, the choice of state-space to regress to drastically affected the performance of the s-map asym-DDPG baseline. In particular, we observed that if we kept the regressor state as quaternions (for Jaco arm links; this is our default state-space setup), that the performance was considerably worse than regressing to cartesian positions and rotations, and significantly worse than simply regressing to cartesian positions (see Figure \ref{fig:s-map_ablation}). Figure \ref{fig:hold_out_envs_rotation} demonstrates that it is the inability to accurately regress to quaternions and cartesian rotations that leads to inferior policy performance for these two s-map asym-DDPG ablations. \citet{zhou2019continuity} similarly observed that quaternions are hard for neural networks to regress and showed that it was due to their representations being discontinuous. It is for this reason why regressing \textbf{only} to cartesian positions performed best.

However, even with a representation which is better suited for learning, the agent's performance is still significantly below \met~(see Figure \ref{fig:s-map_ablation}). Given that the state-space agent used under the \met~framework learns efficiently for this domain, this suggests that the remainder of the s-map asymmetric DDPG policy (layers dependent on the state-space predictor) is rather sensitive to inaccuracies in the regressor. Different methods for using privileged information, as given by \met's attention mechanism, provide more robust performance. 

\end{document}